\documentclass{article}
\usepackage{graphicx}
\usepackage{listings}
\usepackage{enumitem}
\usepackage{hyperref}
\hypersetup{colorlinks=true, linkcolor=blue, urlcolor=blue}
\usepackage{xcolor}
\usepackage{amsmath}  
\usepackage{float}
\usepackage{amsmath}    
\usepackage{amssymb}    
\usepackage{hyperref}   

\newtheorem{theorem}{Theorem}
\usepackage{arxiv}

\usepackage[utf8]{inputenc} 
\usepackage[T1]{fontenc}    
\usepackage{hyperref}       
\usepackage{url}            
\usepackage{booktabs}       
\usepackage{amsfonts}       
\usepackage{nicefrac}       
\usepackage{microtype}      
\usepackage{lipsum}
\usepackage{graphicx}
\graphicspath{ {./images/} }
\usepackage{geometry}
\usepackage{multirow,booktabs}
\usepackage{enumitem}
\setlist[itemize]{nosep,leftmargin=*}

\title{GTS Forecaster: a novel deep learning based geodetic time series forecasting toolbox with python}

\author{
  Xuechen Liang\textsuperscript{1,7}, Xiaoxing He\textsuperscript{2,7,*}, Shengdao Wang\textsuperscript{3,7}, Jean-Philippe Montillet\textsuperscript{4}, \\
  Zhengkai Huang\textsuperscript{1}, Gaël Kermarrec\textsuperscript{5}, Shunqiang Hu\textsuperscript{6}, Yu Zhou\textsuperscript{2}, Jiahui Huang\textsuperscript{2} \\
  \\
  \textsuperscript{1} School of Transportation Engineering, East China Jiao Tong University, Nanchang, China. \\
  \textsuperscript{2} Jiangxi Provincial Key Laboratory of Water Ecological Conservation in Headwater Regions (2023SSY02031), \\
  Jiangxi University of Science and Technology, Ganzhou, China. \\
  \textsuperscript{3} Division of Geodetic Science, School of Earth Sciences, The Ohio State University, Columbus, Ohio 43210, USA. \\
  \textsuperscript{4} Institute Dom Luiz, University of Beira Interior, Covilhã, Portugal. \\
  \textsuperscript{5} Institute of Meteorology and Climatology, Leibniz University Hannover, Herrenhäuserstr, Hannover, Germany. \\
  \textsuperscript{6} Key Laboratory of Poyang Lake Wetland and Watershed Research, Ministry of Education, \\
  Jiangxi Normal University, Nanchang, China. \\
  \textsuperscript{7} These authors contributed equally. \\
  \\
  \texttt{*xxh@jxust.edu.cn} \\
}

\begin{document}
\maketitle
\begin{abstract}
Geodetic time series—such as Global Navigation Satellite System (GNSS) positions, satellite altimetry-derived sea surface height (SSH), and tide gauge (TG) records—is essential for monitoring surface deformation and sea level change. Accurate forecasts of these variables can enhance early warning systems and support hazard mitigation for earthquakes, landslides, coastal storm surge, and long-term sea level. However, the nonlinear, non-stationary, and incomplete nature of such variables presents significant challenges for classic models, which often fail to capture long-term dependencies and complex spatiotemporal dynamics. We introduce GTS Forecaster, an open-source Python package for geodetic time series forecasting. It integrates advanced deep learning models—including kernel attention networks (KAN), graph neural network-based gated recurrent units (GNNGRU), and time-aware graph neural networks (TimeGNN)—to effectively model nonlinear spatial-temporal patterns. The package also provides robust preprocessing tools, including outlier detection and a reinforcement learning-based gap-filling algorithm, the Kalman-TransFusion Interpolation Framework (KTIF). GTS Forecaster currently supports forecasting, visualization, and evaluation of GNSS, SSH, and TG datasets, and is adaptable to general time series applications. By combining cutting-edge models with an accessible interface, it facilitates the application of deep learning in geodetic forecasting tasks.
\end{abstract}


\section{Introduction}
Geodetic time series—including GNSS position data, tide gauge (TG) records, and satellite altimetry-derived sea surface height (SSH)—are critical for understanding Earth’s dynamic processes\cite{he2017review}. These data are extensively applied to monitor surface deformation and sea level change \cite{blewitt2016midas}. Reliable forecasting of such time series plays a vital role in hazard mitigation and long-term planning \cite{fialko2006interseismic}\cite{avouac2015geodetic}. However, geodetic time series exhibit nonlinear and non-stationary behaviors due to geophysical, climatic, and anthropogenic influences, posing significant challenges for accurate modeling and forecasting \cite{montillet2019geodetic}\cite{rebischung2016igs}\cite{he2017review}. Classical statistical models (e.g., Autoregressive integrated moving average) and machine learning methods (e.g., support vector machine) struggle to model geodetic time series due to their reliance on linear assumptions and limited capacity for capturing non-stationary, multi-scale spatiotemporal dependencies., limiting predictive accuracy \cite{box2015time}\cite{hyndman2018forecasting}. For example, Autoregressive integrated moving average (ARIMA) models\cite{box2015time} and standard machine learning techniques \cite{hyndman2018forecasting} have demonstrated utility in stationary scenarios; their limited capacity to capture multi-scale temporal dependencies often leads to degraded predictive performance in real-world geodesy applications.
Recent advances in deep learning have demonstrated substantial potential for time series forecasting across a wide range of scientific disciplines \cite{lecun2015deep}\cite{goodfellow2016deep}\cite{li2025modeling}. Neural networks, particularly in recurrent and graph-based architectures, excel in detecting hidden patterns, capturing long-range dependencies, and adapting to novel scenarios . Nevertheless, the application of deep learning techniques in geodetic forecasting remains relatively limited, largely due to the absence of accessible, well-integrated toolkits tailored to the specific needs of geoscientific research.
To address this issue, we present GTS-Forecaster software, a Python-based toolkit designed for geodetic time series forecasting. The toolkit integrates diverse deep learning models, including Long Short-Term Memory (LSTM), Gated Recurrent Unit (GRU), Temporal Convolutional Network (TCN), Bidirectional LSTM (BiLSTM), Transformer, and Informer, as well as advanced models such as Kolmogorov–Arnold Networks\cite{liu2024kan}, graph neural network-enhanced GRUs (GNN-GRU), and time-aware graph neural networks\cite{xu2023timegnn} . This model suite supports robust and adaptable forecasting for various geodetic time series applications. In addition, the GTS-Forecaster software also provides preprocessing tools such as outlier detection and a novel reinforcement learning-based graph-informed autoregressive (KTIF) gap-filling algorithm. Furthermore, a weighted quality evaluation (WQE) index is proposed to quantitatively evaluate forecasting accuracy, facilitating effective model selection \cite{zhou2025improved}.

\section{Software platform}
\label{sec:headings}
\subsubsection{Software installation}
GTS Forecaster is an open-source software developed using Python and available for download at https://github.com/heimy2000/GTS\_Forecaster. The downloadable package includes installation files, code files, sample data, user manual. Following the download, users can run GTS\_Forecaster toolbox through the following steps:
\subsubsection{Step 1: Obtain Source Code}
Cloning a repository of GTS Forecaster to a local computer.

\begin{lstlisting}
# Clone the official GTS_Forecaster repository
git clone https://github.com/heimy2000/GTS_Forecaster.git

# Navigate to the project root directory
cd GTS_Forecaster
\end{lstlisting}

User can directly download the Source code to replace this step as an ‌alternation‌. 

\subsubsection{Step 2: Setup a conda environment }

\begin{lstlisting}
# Create a Conda environment named "forecast" with Python 3.8
conda create -n forecast python=3.8

# Activate the created environment
conda activate forecast
\end{lstlisting}

\subsubsection{Step 3: Install external dependencies}

\begin{lstlisting}
# Install dependencies from requirements.txt
pip install -r requirements.txt
\end{lstlisting}

\subsubsection{Step 4:GUI}
GTS\_Forecaster provides a user-friendly GUI developed with \textbf{Tkinter}, enabling time series forecasting through intuitive clicks (no complex coding required). Key features of the GUI include:
\begin{figure}[H]  
    \centering
    \includegraphics[width=0.8\textwidth]{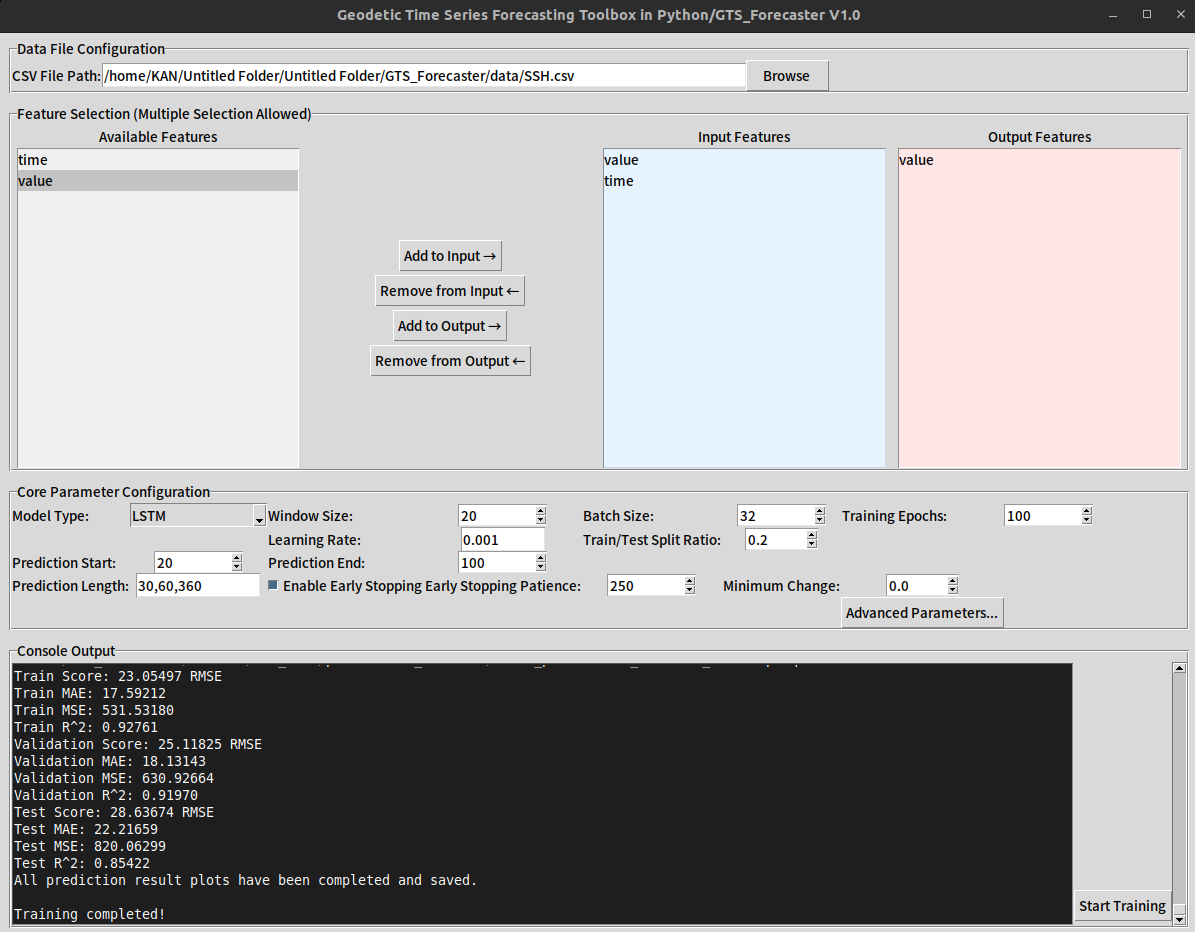} 
    \caption{Primary Interface of GTS\_Forecaster Software}
    \label{fig:gui_main}
\end{figure}

\subsection{Features of GTS Forecaster}
GTS Forecaster offers functionalities grouped in three main modules: the geodetic time series preprocessing module, the geodetic time series forecasting module, foand the recasting results visualization and accuracy evaluation module. Fig. 2 shows a schematic diagram of the architecture of the GTS Forecaster software. More details can also be found in the user manual.

\begin{figure}[H]
    \centering  
    \includegraphics[width=0.8\textwidth]{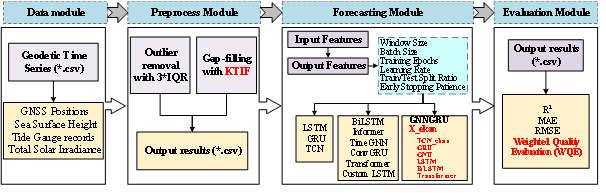}
    \caption{The structure diagram of GTS Forecaster. }
    \label{fig:gts_gui}  

\end{figure}

\section{Geodetic Time Series Preprocessing}
The geodetic time series preprocessing includes the removal of outliers and handling of data gaps. 

\subsection{Outlier removal}
In geodetic time series forecasting, the outlier removal phase is a crucial pre-step reprocessing. Outliers introduce noise, interfere with the model’s ability to learn the true signal, and may lead to overfitting or unstable training processes \cite{ackerman2020detection}. Additionally, outliers can distort statistical metrics, affect data normalization, and cause biases or abrupt predictions that fail to reflect the smooth continuity of geodetic processes, such as crustal movement or deformation [REF?]. By eliminating outliers, the input data quality is enhanced, ensuring the model focuses on true physical processes (e.g., tectonic rate), therefore generating more accurate and reliable forecasting results. GTS Forecaster relies on the standard used of the 3 interquartile range (3IQR) algorithm \cite{seo2006review}. Any users can run the outlier removal function with the routine “python outlier 3IQR.py”. Fig. 3 displays the three coordinates (ENU) of a daily position GNSS time series for the station AC03 (as an example). 

\begin{figure}[H]
    \centering  
    \includegraphics[width=0.8\textwidth]{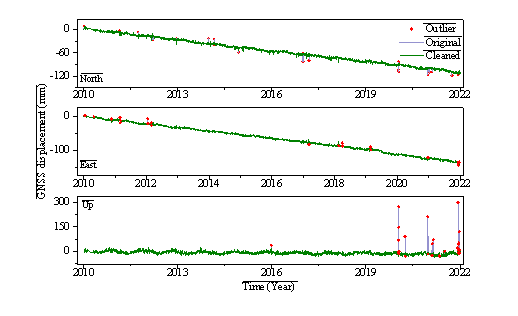}
    \caption{Example of GNSS time series (AC03) outlier detection with 3IQR }
    \label{fig:gts_gui}  

\end{figure}

\subsection{Geodetic time series Gap-filling with KTIF algorithm}

The KTIF framework addresses the unique challenge of geodetic data gaps by fusing geophysical priors (e.g., plate motion continuity) with data-driven learning, surpassing traditional interpolation methods that neglect spatial-temporal correlations in crustal deformation. It begins with Dynamic System Parametric Modeling (DSPM) to construct hidden state-space representations that encode crustal deformation and tidal dynamics via differential equation constraints. Missing data intervals are addressed through Smoothing Estimation with Missing Data, which employs bidirectional Kalman recursions. Forward filtering predicts system evolution from historical observations, while backward smoothing (using the Rauch-Tung-Striebel algorithm) refines estimates using future information for spatiotemporally continuous interpolation. To ensure physical fidelity, a Dynamic Consistency Constraint mechanism quantifies deviations between interpolated trajectories and PDE-governed system behaviors using Wasserstein distance metrics, dynamically adjusted by spatiotemporal attention networks to balance noise suppression and plate motion preservation across stations. The Parametric Inverse Correction Process completes the workflow through reversible neural operators that generate backward-propagated state sequences, fused with forward estimates via Jacobian-constrained gating networks to optimize the synergy between physical priors and learned features
To realize the transformation of theoretical frameworks into computable algorithms, the KTIF method employs a four-stage progressive modeling paradigm that systematically embeds geophysical dynamics priors and data-driven mechanisms into computational workflows. Its core lies in constructing a bidirectional closed-loop optimization system.

\subsubsection{Step 1: Dynamic System Parametric Modeling}

The Dynamic System Parametric Modeling (DPSM) establishes a hidden Markov state-space representation with latent variables and noise covariance.
State Space Representation
Recent studies combining experimental observations with theoretical analyses \cite{arismendi2021piecewise} have revealed that the dynamic evolution of GNSS/TG observation sequences inherently exhibits hidden Markov properties. To establish a unified characterization framework that systematically integrates the underlying physical principles with distinctive data features of these systems, we propose a parametric tuple defined as follows \cite{xie2025stability}:

\begin{equation}
\left\{
\begin{aligned}
x_t &= F x_{t-1} + w_t, & w_t &\sim \mathcal{N}(0, Q), \\
z_t &= H x_t + v_t,   & v_t &\sim \mathcal{N}(0, R).
\end{aligned}
\right.
\end{equation}

$x_t \in \mathbb{R}^n$ represents the latent state variable. 

$F \in \mathbb{R}^{n \times n}$ is the state transition matrix.

\( w_t \) denotes process noise.

\( v_t \) represents Gaussian-distributed observation noise.

\( z_t \) is the observation at time \( t \).

\( H \) is the observation matrix.

\( Q \) and \( R \) are the covariance matrices of process noise and observation noise.
\subsubsection{Step 2: Smoothing Estimation with Missing Data}
\begin{equation}
x_t^s = x_t^f + P_t^f F^T \left( F P_t^f F^T + Q \right)^{-1} \left( x_{t+1}^s - F x_t^f \right)
\end{equation}

\( P_t^f \) represents the forward covariance matrix, which is used to control the reliability of the interpolation .

\subsubsection{Step 3: Dynamic Consistency Constraint}
\begin{equation}
\| f_\theta(z_{1:t-1}) - H F \widehat{x}_{t-1} \|_2^2 \leq \chi_{d,\alpha}^2
\end{equation}

Left-hand side: The squared Mahalanobis distance between the interpolation result $f_\theta(z_{1:t-1})$ and the system prediction $H F \widehat{x}_{t-1}$.

Right-hand side: The threshold $\chi_{d,\alpha}^2$ for the rejection region in hypothesis testing, corresponding to the 95\% confidence interval when $\alpha$=0.05.

\begin{theorem}[Attention-Augmented Interpolation Operator]
Construct a spatiotemporal attention weight matrix \( A \in \mathbb{R}^{T \times T} \):
\begin{equation}
A_{ij} = \frac{\exp\left( \phi(q_i, k_j) + \log p_F(|i-j|) \right)}{\sum_{k=1}^T \exp\left( \phi(q_i, k_k) + \log p_F(|i-k|) \right)}
\end{equation}
where:
\begin{itemize}
    \item \( \phi(q_i, k_j) \) is the Query-Key similarity function.
    \item \( p_F(\Delta t) = \exp\left( -\lambda \| F^{\Delta t} - I \|_F \right) \) is the temporal decay term based on the state transition matrix.
\end{itemize}
The imputed value for the missing time instant \( t \in \Omega_{\text{miss}} \) is:
\begin{equation}
\widehat{z}_t = \underbrace{H \widehat{x}_t^s}_{\text{smoothing baseline}} + \underbrace{\sum_{j \in \Gamma(t)} A_{tj} \left( z_j - H \widehat{x}_j^s \right)}_{\text{Contextual correction term}}
\end{equation}
where \( \Gamma(t) \) represents the spatiotemporal neighborhood window centered at \( t \) .
\end{theorem}

\subsubsection{Step 4: Parametric Inverse Correction Process}

\subsubsubsection{Inverse Dynamics Operator}
To enhance the robustness of state estimation during missing time intervals, the inverse state transition equation is established:
\begin{equation}
x_{t-1}^\flat = F^{-1} x_t^\flat + \xi_t, \quad \xi_t \sim \mathcal{N}\!\left(0, F^{-1} Q (F^{-1})^\top \right)
\end{equation}

Dual-channel features are constructed using the forward smoothing sequence $\{ x_t^s \}$ and the inverse estimation $\{ x_t^\flat \}$:

\begin{itemize}
    \item $x_{t-1}^\flat$ represents the state of the inverse estimation system at time $t\!-\!1$.
    \item $F^{-1}$ is the inverse of the state transition matrix $F$.
    \item $\xi_t$ is the inverse process noise.
\end{itemize}

\subsubsubsection{Feature Fusion Equation}
\begin{equation}
\widehat{z}_t = \gamma_t \cdot \operatorname{MLP}(x_t^s) + (1 - \gamma_t) \cdot \operatorname{Attn}(x_{1:t}^\flat)
\end{equation}

\textbf{Components:}
\begin{itemize}[leftmargin=*]
    \item $\widehat{z}_t$: Final interpolated result at time $t$.
    \item \textbf{Gating Mechanism}: \\
          $\gamma_t = \sigma\!\left(w^\top [x_t^s \, x_t^\flat]\right)$ adaptively assigns weights to the forward and inverse features.
    \item $\gamma_t$: Gating coefficient balancing the contributions of the forward smoothing feature $\operatorname{MLP}(x_t^s)$ and the inverse contextual feature $\operatorname{Attn}(x_{1:t}^\flat)$.
    \begin{itemize}
        \item $\sigma(\cdot)$: Sigmoid function ensuring bounded weights.
        \item $w$: Learnable weight vector for feature combination.
    \end{itemize}
    \item $[x_t^s \, x_t^\flat]$: Concatenation of the forward-smoothed state $x_t^s$ and inverse-estimated state $x_t^\flat$.
\end{itemize}

Fig. 4 shows the interpolation performance of the KITF algorithm on GNSS/TG data. Based on the original low missing rate data, after randomly deleting 5\% of the observed values, the goodness of fit ($R^2$) between the algorithm’s interpolation results and the true values reached 0.95 (GNSS) and 0.99 (TG), respectively—verifying its high-precision recovery ability in sparse missing scenarios.

\begin{figure}[H]
    \centering  
    \includegraphics[width=0.8\textwidth]{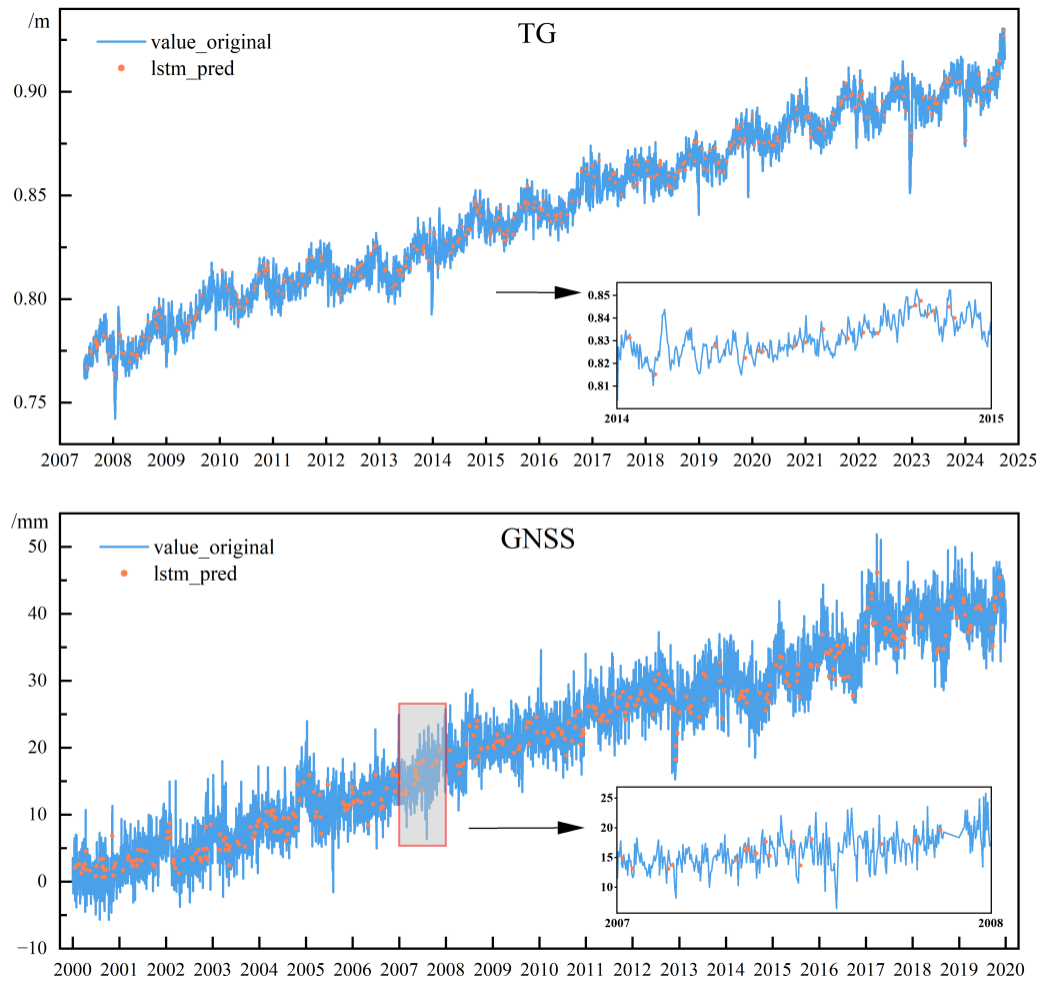}
    \caption{The above figure shows the interpolation effect of the KTIF algorithm. }
    \label{fig:gts_gui}  

\end{figure}

\subsection{Geodetic time series forecasting with deep learning}

After completing the preprocessing of geodetic time series data, including outlier removal and gapfilling, the next step is to conduct forecasting using advanced deep learning models, which is the core functionality of GTS\_Forecaster. GTS\_Forecaster integrates three categories of deep learning architectures for geodetic time series forecasting: baseline models (established architectures), enhanced variants (modified versions of baseline models), and novel hybrid algorithms (proposed in this work). These models address distinct challenges in geodetic data, such as nonlinearity, non-stationarity, and spatiotemporal dependencies. A summary of available models and their characteristics is provided in Table 1.

\begin{table}[htbp]
  \small
  \centering
  \caption{Comparison of Algorithms, Prediction Models and Their Features}
  \label{tab:full}
  \begin{tabular}{@{}llp{9.6cm}@{}}
    \toprule
    \textbf{Algorithms} & \textbf{Prediction Models} & \textbf{Features of Available Models} \\
    \midrule
    
    \multirow{3}{*}{Prototype Algorithms}
    & LSTM \cite{hochreiter1997long}
    & \begin{itemize}
        \item \textbf{Strengths:} Solves long-term dependency issues via gating mechanisms.
        \item \textbf{Weaknesses:} High computational complexity and low parallelism.
      \end{itemize} \\
    \cmidrule(l){2-3}
    & GRU \cite{cho2014learning}
    & \begin{itemize}
        \item \textbf{Strengths:} Parameter efficiency and fast training.
        \item \textbf{Weaknesses:} Weaker long-sequence modeling than LSTM.
      \end{itemize} \\
    \cmidrule(l){2-3}
    & TCN \cite{bai2018empirical}
    & \begin{itemize}
        \item \textbf{Strengths:} Parallel computation and stable gradients.
        \item \textbf{Weaknesses:} Sensitive to input length; limited local feature capture.
      \end{itemize} \\
    \midrule
    
    \multirow{6}{*}{Modified Algorithms}
    & Custom LSTM \cite{girshick2015fast}
    & \begin{itemize}
        \item \textbf{Strengths:} High precision, end-to-end training, multi-task learning.
        \item \textbf{Weaknesses:} Computationally heavy; limited small-object detection; relies on region proposals.
      \end{itemize} \\
    \cmidrule(l){2-3}
    & Transformer 
    & \begin{itemize}
        \item \textbf{Strengths:} Global dependency modeling; highly parallel.
        \item \textbf{Weaknesses:} Quadratic complexity; depends on positional encoding.
      \end{itemize} \\
    \cmidrule(l){2-3}
    & ConvGRU \cite{shi2017deep}
    & \begin{itemize}
        \item \textbf{Strengths:} Spatiotemporal feature fusion.
        \item \textbf{Weaknesses:} Convolutional kernels limit long-range dependencies.
      \end{itemize} \\
    \cmidrule(l){2-3}
    & BiLSTM \cite{huang2015bidirectional}
    & \begin{itemize}
        \item \textbf{Strengths:} Uses past and future context; expressive for long sequences.
        \item \textbf{Weaknesses:} Computationally intensive; prone to overfitting; long training time.
      \end{itemize} \\
    \cmidrule(l){2-3}
    & Informer \cite{zhou2021informer}
    & \begin{itemize}
        \item \textbf{Strengths:} ProbSparse attention yields linear complexity on long series.
        \item \textbf{Weaknesses:} Requires careful hyper-parameter tuning.
      \end{itemize} \\
    \cmidrule(l){2-3}
    & TimeGNN
    & \begin{itemize}
        \item \textbf{Strengths:} Dynamic spatiotemporal graph modeling.
        \item \textbf{Weaknesses:} Limited causal interpretability; high preprocessing cost.
      \end{itemize} \\
    \midrule
    
    \multirow{2}{*}{Proposed Novel Algorithms}
    & X-eKAN 
    & \begin{itemize}
        \item \textbf{Strengths:} Strong interpretability; embeds scientific laws; high accuracy on specific tasks; parameter-efficient.
        \item \textbf{Weaknesses:} Slow training; instability; high compute and memory cost.
      \end{itemize} \\
    \cmidrule(l){2-3}
    & GRUGNN
    & \begin{itemize}
        \item \textbf{Strengths:} Fuses spatial and temporal features via GNN and GRU.
        \item \textbf{Weaknesses:} Struggles with long-range dependencies; limited explainability.
      \end{itemize} \\
    \bottomrule
  \end{tabular}
\end{table}

\subsection{Proposed Novel Algorithms eKAN}

The Kolmogorov-Arnold Representation Theorem states that any \( n \)-dimensional continuous function can be precisely represented as a combination of univariate functions. Formulated by Andrey Kolmogorov and Vladimir Arnold, the Kolmogorov–Arnold Representation Theorem establishes that any continuous multivariate function can be expressed as a finite superposition of univariate functions. It asserts that any continuous multivariate functions can be decomposed into a series of simple functions, formally expressed as nested univariate functions \cite{liu2024kan}.

Following this theory, the Kolmogorov-Arnold Network (KAN) realizes the approximation of complex functions through two key steps. The first step is the coordinate transformation, where each input dimension \( x_p \) is nonlinearly transformed through a learnable univariate function \( h_{q,p} \). The second step is the channel aggregation, where the transformed results are combined through a summation layer and then mapped to the output via a univariate function \( \Phi_q \). This structure decomposes the learning process of high-dimensional functions into the collaborative optimization of univariate functions. Compared to traditional neural networks, KAN has two major advantages. First, since the network depth is independent of the input dimension, the model is more compact, as indicated by Formula (8), representing the mathematical expression of KAN. As a result, the parameter efficiency is greatly improved. Second, the fully nonlinear path can capture more complex functional relationships, thereby enhancing the expressive power, as shown in the expansion of Formula (9)

\begin{equation}
f(x) = \sum_{i=1}^{n} \phi_i(x_i)
\end{equation}

\begin{equation}
g(x) = \Phi\left(\sum_{j=1}^{m} \psi_j(x_j)\right)
\end{equation}

\subsection{Structure and Characteristics of KAN}

The innovation of KAN resides in its unique architectural design, which fundamentally diverges from conventional neural networks through the systematic integration of learnable activation functions into the model's architecture. Unlike traditional approaches that fix activation functions as static nonlinear components, KAN employs trainable activation functions at each network connection. This design paradigm endows every connection with dual adaptive capabilities: conventional weight parameters and trainable nonlinear transformations. Such configuration enables the network to autonomously discover optimal nonlinear mappings during the learning process, significantly enhancing its capacity to model complex functional relationships. The adaptive activation mechanism provides exceptional flexibility in addressing nonlinear challenges, particularly demonstrating superior performance in scenarios involving intricate data distributions or highly nonlinear correlations.

\figurename~5 elucidates the operational mechanism of KAN through a dual-panel illustration. The left panel depicts the propagation pathways of activation signals across network layers, while the right panel provides a detailed exposition of the dynamic modulation principles governing the B-spline-based activation functions. These activation functions are constructed through the superposition of adaptable B-spline basis functions, where the contribution weights of individual basis functions are dynamically optimized during the training process.

\begin{figure}[H]
    \centering  
    \includegraphics[width=0.8\textwidth]{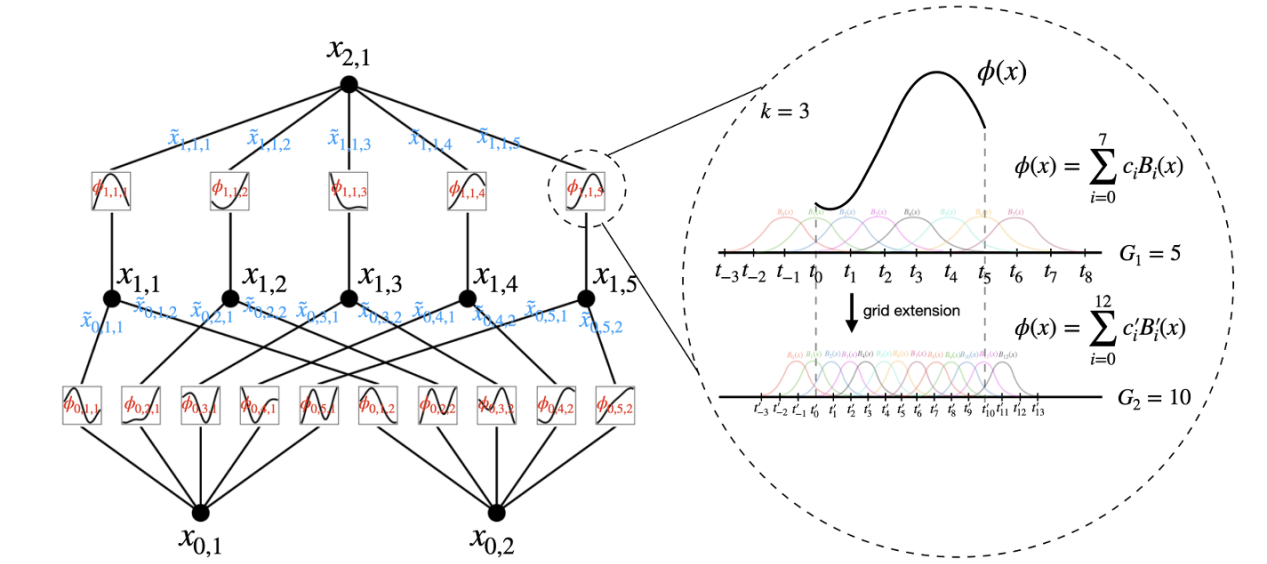}
    \caption{Basic flow of KAN algorithm. }
    \label{fig:gts_gui}  

\end{figure}

\subsection{Proposed Novel Algorithms GRUGNN}
In complex spatiotemporal sequence modeling tasks, combining graph neural networks (GNNs) with gated recurrent units (GRUs) offers an effective solution for analyzing and predicting dynamic graph data. GNNs capture spatial topological dependencies by aggregating neighborhood information. GRUs model temporal dynamic patterns (like event sequences and sensor signals) with gating mechanisms. By inputting GNN - extracted spatial features into GRUs for temporal encoding, or dynamically updating graph structures with GRU hidden states, the combined model can capture the coupling effects of spatiotemporal features \cite{rashnu2024integrative}.

\section{Forecasting results and accuracy evaluation with GTS forcaster}
After employing deep learning models for geodetic time series forecasting, it is essential to assess the accuracy of the results and visualize them for enhanced understanding. To evaluate the model's multi-scale forecasting capabilities, the GNSS/TG/SSH dataset was subjected to temporal partitioning with strict chronological segregation: 80\% of the time series was designated for the training-validation set (including internal splits for parameter training and hyperparameter optimization/early stopping), while the remaining 20\% was reserved as an independent test set. For example, in a temporal sequence within 100 days, the initial 80 days constituted the combined training-validation data, with the final 20 days serving as a completely unseen test partition. Three distinct forecasting tasks were established: short-term forecasting (Y1) aimed at capturing transient patterns within the first 1\% of the test period; medium-term forecasting (Y2) covering 50\% of the test duration to evaluate periodic behavior, and long-term forecasting (Y3) encompassing the entire test horizon to evaluate systematic trends. 

Tab. 2~4 present comparisons of forecasting results for different deep learning models applied to TG, GNSS, and SSH time series. For TG time series forecasting, the LSTM\_ekan model outperforms other models at short, medium, and long-term predictions, and the WQE value increased by 44\%, 27\%, 25\% and 35.1\%, 35.1\%, 52.8\% in comparison of LSTM and GRU model, respectively. For GNSS time series forecasting, the BiLSTM\_ekan model shows more performance compared with other models at medium, and long-term predictions, with WQE value increased by 10\% over LSTM model, whereas the LSTM model excels in short predictions. Additionally, the WQE value with BiLSTM\_ekan model increased by 23\%, 23\%, 24\% at short, medium, and long-term predictions in comparison of TimeGNN model. For SSH time series forecasting, the BiLSTM\_ekan model shows more performance compared with other models at short, medium, and long-term predictions, with WQE value increased by 15\%, 15\%, 23\% and 21\%, 23\%, 22\% compared with LSTM and BiLSTM model, respectively. Furthermore, Figure 6-8 visually compare the forecasting accuracy of LSTM\_ekan, LSTM, GRU and Transformer models against true values in the test set.

In summary, the models optimized by KAN (e.g., LSTM-EKAN, BiLSTM-EKAN), with complex architecture, effectively capture long-term dependencies and intricate patterns in long-term forecasting tasks, outperforming MLP model. However, 
MLP model, with streamlined structure, efficiently learn key features in short-term forecasting tasks, yielding superior results. the forecasting results in this study demonstrate that the models optimized by KAN can significantly improve WQE values in long-term forecasting, validating their capacity to model temporal dependencies. Meanwhile, MLP models (e.g., LSTM, BiLSTM) excel in short-term forecasting due to their rapid feature-learning capabilities.These findings align with theoretical expectations, confirming the comparative advantages of KAN-optimized models and MLP in respective forecasting tasks.

\begin{table}[H]
  \caption{Comparative Analysis of TG Time Series Forecasting Results Across Different Models (Short, Medium, Long Term)}
  \centering
  \begin{tabular}{lrrrrrrrrrrrr}  
    \toprule
    \multirow{2}{*}{Tide Gauge} & \multicolumn{4}{c}{Long-20\%} & \multicolumn{4}{c}{Mid-10\%} & \multicolumn{4}{c}{Short-1\%} \\
    \cmidrule(r){2-5} \cmidrule(r){6-9} \cmidrule(r){10-13}  
    & R² & RMSE & MAE & WQE & R² & RMSE & MAE & WQE & R² & RMSE & MAE & WQE \\
    \midrule
    LSTM & 0.84 & 29.91 & 23.22 & 0.85 & 0.84 & 29.63 & 23.08 & 0.84 & 0.89 & 24.71 & 18.8 & 0.61 \\
    LSTM\_ekan & 0.88 & 18.35 & 14.04 & 0.61 & 0.88 & 18.4 & 14.02 & 0.61 & 0.94 & 14.7 & 11.3 & 0.34 \\
    GRU & 0.82 & 31.73 & 24.92 & 0.94 & 0.82 & 31.5 & 24.71 & 0.94 & 0.87 & 27.86 & 22.5 & 0.72 \\
    TCN & 0.85 & 28.62 & 22.19 & 0.8 & 0.86 & 28.05 & 21.96 & 0.75 & 0.9 & 24.64 & 19.57 & 0.57 \\
    TCN\_ekan & 0.86 & 27.92 & 20.41 & 0.75 & 0.86 & 27.77 & 20.59 & 0.75 & 0.9 & 24.02 & 19.3 & 0.56 \\
    BiLSTM & 0.83 & 30.11 & 22.62 & 0.88 & 0.84 & 29.8 & 22.68 & 0.84 & 0.88 & 25.9 & 19.99 & 0.65 \\
    BiLSTM\_ekan & 0.84 & 30.24 & 23.62 & 0.84 & 0.84 & 29.77 & 23.16 & 0.84 & 0.89 & 25.4 & 20.56 & 0.61 \\
    Informer & 0.84 & 30.27 & 22.78 & 0.84 & 0.84 & 29.55 & 22.72 & 0.84 & 0.88 & 26.47 & 20.83 & 0.66 \\
    GNNGRU & 0.84 & 29.69 & 23.54 & 0.84 & 0.85 & 29.56 & 23.51 & 0.8 & 0.89 & 26.07 & 20.58 & 0.62 \\
    CustomLSTM & 0.85 & 29.33 & 22.03 & 0.8 & 0.85 & 29.41 & 22.05 & 0.8 & 0.89 & 25.54 & 19.52 & 0.61 \\
    Transformer & 0.82 & 31.89 & 24.43 & 0.94 & 0.82 & 31.93 & 24.5 & 0.94 & 0.87 & 27.79 & 22.27 & 0.72 \\
    Transformer\_ekan & 0.87 & 26.72 & 19.66 & 0.7 & 0.88 & 26.14 & 19.5 & 0.66 & 0.92 & 22.36 & 16.84 & 0.47 \\
    ConvGRU & 0.84 & 30.19 & 22.24 & 0.84 & 0.84 & 29.66 & 22.09 & 0.84 & 0.89 & 25.97 & 19.87 & 0.62 \\
    TimeGNN & 0.84 & 31.01 & 24.66 & 0.62 & 0.85 & 28.84 & 21.98 & 0.86 & 0.89 & 25.48 & 19.58 & 0.80 \\
    \bottomrule
  \end{tabular}
  \label{tab:tg_forecasting_results}
\end{table}
For the prediction of TG sea level height sequences, LSTM\_ekan demonstrated optimal performance in short-, medium-, and long-term predictions of tide gauge station sequences. Compared to the native LSTM model, the WQE improved by (0.61-0.34)/0.61=44\%, (0.61-0.84)/0.84=27\%, and (0.61-0.85)/0.85=25\%, respectively. Additionally, LSTM\_ekan achieved WQE improvements of 35.1\%, 35.1\%, and 52.8\% over the GRU model. To provide a more intuitive comparison of the prediction results across different models, Figure 6 presents a comparison of LSTM\_ekan, LSTM, GRU, and Transformer model fits against true values on the test set (the zoomed-in panel highlights the period of the annotated window) for the TG time series.

\begin{figure}[H]
    \centering  
    \includegraphics[width=0.8\textwidth]{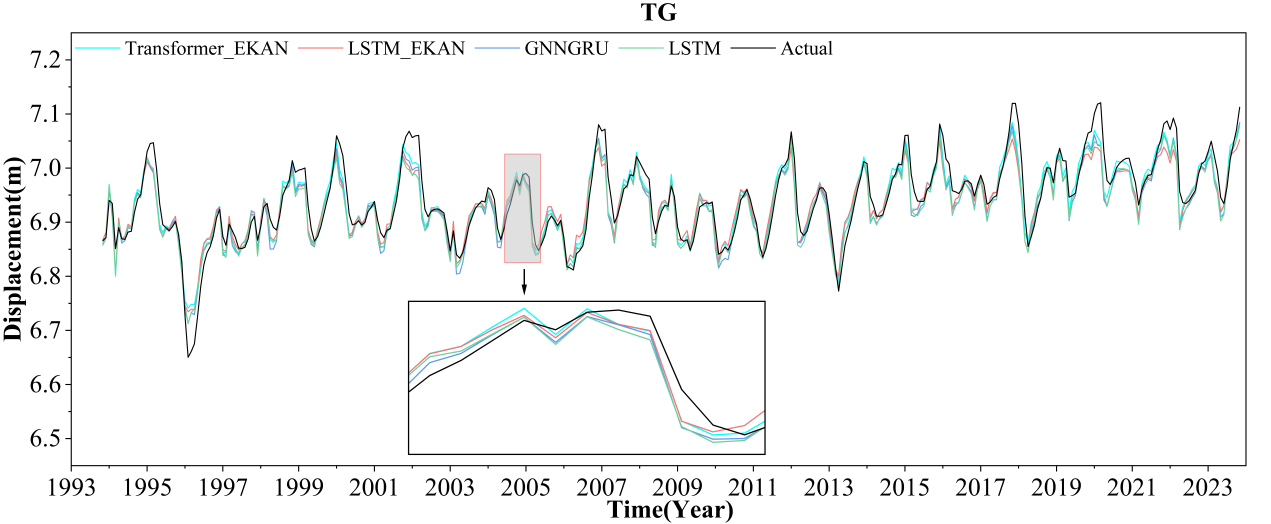}
    \caption{Comparison of LSTM\_ekan, LSTM, GRU and Transformer model fits against true values on the test set (the zoomed-in panel highlights the period of the annotated window) on TG time series}
    \label{fig:gts_gui}  

\end{figure}

\begin{table}[H]
  \centering
  \caption{Forecasting Results of Different Deep Learning Models for Tide Gauge Time Series}
  \label{tab:tg_forecast}
  \begin{tabular}{lrrrrrrrrrrrr}
    \toprule
    \multirow{2}{*}{\textbf{Tide Gauge}} & \multicolumn{4}{c}{\textbf{Long-20\%}} & \multicolumn{4}{c}{\textbf{Mid-10\%}} & \multicolumn{4}{c}{\textbf{Short-1\%}} \\
    \cmidrule(lr){2-5} \cmidrule(lr){6-9} \cmidrule(lr){10-13}  
    & \textbf{R²} & \textbf{RMSE} & \textbf{MAE} & \textbf{WQE} & \textbf{R²} & \textbf{RMSE} & \textbf{MAE} & \textbf{WQE} & \textbf{R²} & \textbf{RMSE} & \textbf{MAE} & \textbf{WQE} \\
    \midrule
    LSTM           & 0.88 & 0.50 & 0.38 & 0.65 & 0.88 & 0.49 & 0.38 & 0.65 & 0.91 & 0.48 & 0.36 & 0.49 \\
    LSTM\_ekan     & 0.88 & 0.50 & 0.39 & 0.65 & 0.88 & 0.50 & 0.39 & 0.65 & 0.90 & 0.50 & 0.39 & 0.55 \\
    GRU            & 0.88 & 0.50 & 0.38 & 0.65 & 0.88 & 0.49 & 0.38 & 0.65 & 0.90 & 0.49 & 0.38 & 0.55 \\
    TCN            & 0.89 & 0.49 & 0.38 & 0.60 & 0.89 & 0.48 & 0.38 & 0.60 & 0.91 & 0.48 & 0.36 & 0.49 \\
    TCN\_ekan      & 0.88 & 0.51 & 0.39 & 0.65 & 0.88 & 0.50 & 0.39 & 0.65 & 0.90 & 0.49 & 0.38 & 0.55 \\
    BiLSTM         & 0.89 & 0.49 & 0.38 & 0.60 & 0.89 & 0.48 & 0.37 & 0.59 & 0.91 & 0.47 & 0.36 & 0.50 \\
    BiLSTM\_ekan   & 0.89 & 0.49 & 0.37 & 0.59 & 0.89 & 0.48 & 0.37 & 0.59 & 0.91 & 0.47 & 0.37 & 0.50 \\
    Informer       & 0.88 & 0.49 & 0.38 & 0.65 & 0.88 & 0.49 & 0.38 & 0.65 & 0.90 & 0.48 & 0.37 & 0.55 \\
    GNNGRU         & 0.88 & 0.50 & 0.39 & 0.65 & 0.88 & 0.50 & 0.39 & 0.65 & 0.90 & 0.50 & 0.39 & 0.55 \\
    CustomLSTM     & 0.87 & 0.52 & 0.40 & 0.70 & 0.87 & 0.52 & 0.40 & 0.70 & 0.89 & 0.51 & 0.40 & 0.60 \\
    Transformer    & 0.89 & 0.49 & 0.38 & 0.60 & 0.89 & 0.48 & 0.37 & 0.60 & 0.91 & 0.47 & 0.36 & 0.50 \\
    Tranformer\_ekan & 0.89 & 0.49 & 0.38 & 0.60 & 0.89 & 0.48 & 0.38 & 0.60 & 0.91 & 0.47 & 0.36 & 0.50 \\
    ConvGRU        & 0.88 & 0.50 & 0.38 & 0.65 & 0.88 & 0.49 & 0.38 & 0.65 & 0.90 & 0.48 & 0.37 & 0.55 \\
    TimeGNN        & 0.86 & 0.54 & 0.41 & 0.76 & 0.86 & 0.54 & 0.42 & 0.76 & 0.88 & 0.55 & 0.42 & 0.66 \\
    \bottomrule
  \end{tabular}
\end{table}

For GNSS prediction, BiLSTM\_ekan performed optimally in medium- and long-term predictions of tide gauge station sequences. Compared to the native LSTM model, the WQE improved by 10\% and 10\%, respectively, while the standard LSTM performed best in short-term prediction. Additionally, BiLSTM\_ekan achieved improvements of 23\%, 23\%, and 24\% over TimeGNN. Furthermore, to provide a more intuitive comparison of the prediction results across different models, Figure 7 presents the comparative results of BiLSTM, LSTM\_ekan, GNNGRU, and LSTM.

\begin{figure}[H]
    \centering  
    \includegraphics[width=0.8\textwidth]{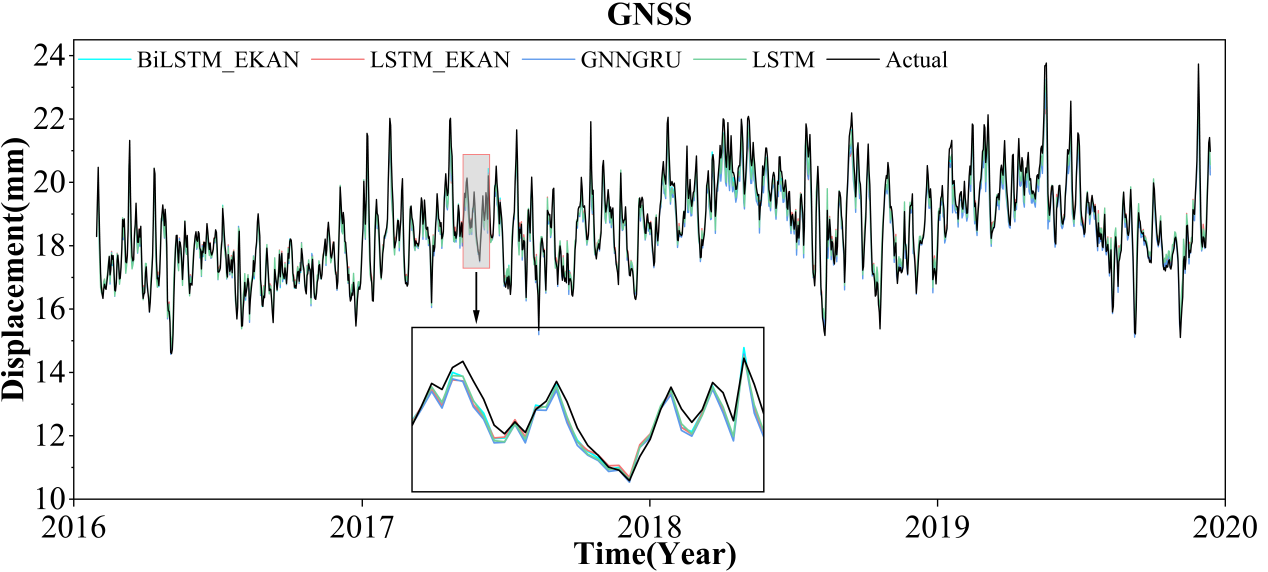}
    \caption{Comparison of LSTM\_ekan, LSTM, GRU and Transformer model fits against true values on the test set (the zoomed-in panel highlights the period of the annotated window) on GNSS time series}
    \label{fig:gts_gui}  

\end{figure}

\begin{table}[H]
  \centering
  \caption{Forecasting Results of Deep Learning Models for Tide Gauge Time Series}
  \label{tab:tide_gauge_forecast}
  \begin{tabular}{lrrrrrrrrrrrr}
    \toprule
    \multirow{2}{*}{\textbf{Tide Gauge}} 
    & \multicolumn{4}{c}{\textbf{Long-20\%}} 
    & \multicolumn{4}{c}{\textbf{Mid-10\%}} 
    & \multicolumn{4}{c}{\textbf{Short-1\%}} \\
    \cmidrule(lr){2-5} \cmidrule(lr){6-9} \cmidrule(lr){10-13}  
    & \textbf{R²} & \textbf{RMSE} & \textbf{MAE} & \textbf{WQE} 
    & \textbf{R²} & \textbf{RMSE} & \textbf{MAE} & \textbf{WQE} 
    & \textbf{R²} & \textbf{RMSE} & \textbf{MAE} & \textbf{WQE} \\
    \midrule
    LSTM           & 0.88 & 18.27 & 13.87 & 0.78 & 0.89 & 17.45 & 13.05 & 0.71 & 0.92 & 16.54 & 13.57 & 0.57 \\
    LSTM\_ekan     & 0.88 & 18.35 & 14.04 & 0.78 & 0.88 & 18.40 & 14.03 & 0.78 & 0.94 & 14.70 & 11.31 & 0.44 \\
    GRU            & 0.89 & 17.37 & 13.17 & 0.72 & 0.90 & 16.77 & 12.47 & 0.66 & 0.94 & 14.33 & 11.14 & 0.44 \\
    TCN            & 0.89 & 17.76 & 13.33 & 0.72 & 0.90 & 16.94 & 12.50 & 0.66 & 0.93 & 15.71 & 12.50 & 0.51 \\
    TCN\_ekan      & 0.88 & 18.67 & 14.43 & 0.78 & 0.88 & 18.06 & 13.96 & 0.78 & 0.91 & 17.18 & 14.32 & 0.63 \\
    BiLSTM         & 0.87 & 19.11 & 15.34 & 0.84 & 0.88 & 18.46 & 14.72 & 0.78 & 0.89 & 18.79 & 16.05 & 0.74 \\
    BiLSTM\_ekan   & 0.90 & 17.01 & 12.83 & 0.66 & 0.91 & 16.16 & 11.98 & 0.60 & 0.94 & 14.36 & 10.94 & 0.44 \\
    Informer       & 0.89 & 17.43 & 13.27 & 0.72 & 0.90 & 16.80 & 12.47 & 0.66 & 0.94 & 14.70 & 10.92 & 0.44 \\
    GNNGRU         & 0.90 & 16.48 & 12.33 & 0.66 & 0.91 & 15.83 & 11.65 & 0.60 & 0.94 & 14.39 & 11.18 & 0.44 \\
    CustomLSTM     & 0.90 & 17.20 & 13.03 & 0.67 & 0.90 & 16.63 & 12.48 & 0.66 & 0.93 & 15.50 & 11.63 & 0.50 \\
    Transformer    & 0.87 & 18.86 & 14.95 & 0.84 & 0.88 & 18.37 & 14.51 & 0.78 & 0.90 & 18.20 & 15.47 & 0.69 \\
    Transformer\_ekan & 0.87 & 19.05 & 15.19 & 0.84 & 0.88 & 18.45 & 14.62 & 0.78 & 0.90 & 18.41 & 15.65 & 0.69 \\
    ConvGRU        & 0.90 & 16.74 & 12.64 & 0.66 & 0.91 & 16.14 & 12.00 & 0.60 & 0.94 & 14.76 & 11.75 & 0.45 \\
    TimeGNN        & 0.88 & 18.16 & 14.40 & 0.72 & 0.89 & 17.46 & 13.70 & 0.66 & 0.91 & 17.24 & 14.61 & 0.57 \\
    \bottomrule
  \end{tabular}
\end{table}
For SSH sequence prediction, BiLSTM\_ekan demonstrated the best performance in short-, medium-, and long-term predictions of tide gauge station sequences. Compared to the native LSTM model, the WQE improved by 15\%, 15\%, and 23\%, respectively. Additionally, BiLSTM\_ekan outperformed BiLSTM, with WQE improvements of 21\%, 23\%, and 22\%. To provide a more intuitive comparison of the prediction results across different models, Figure 8 presents the performance of BiLSTM, LSTM\_ekan, GNNGRU, and LSTM.

\begin{figure}[H]
    \centering  
    \includegraphics[width=0.8\textwidth]{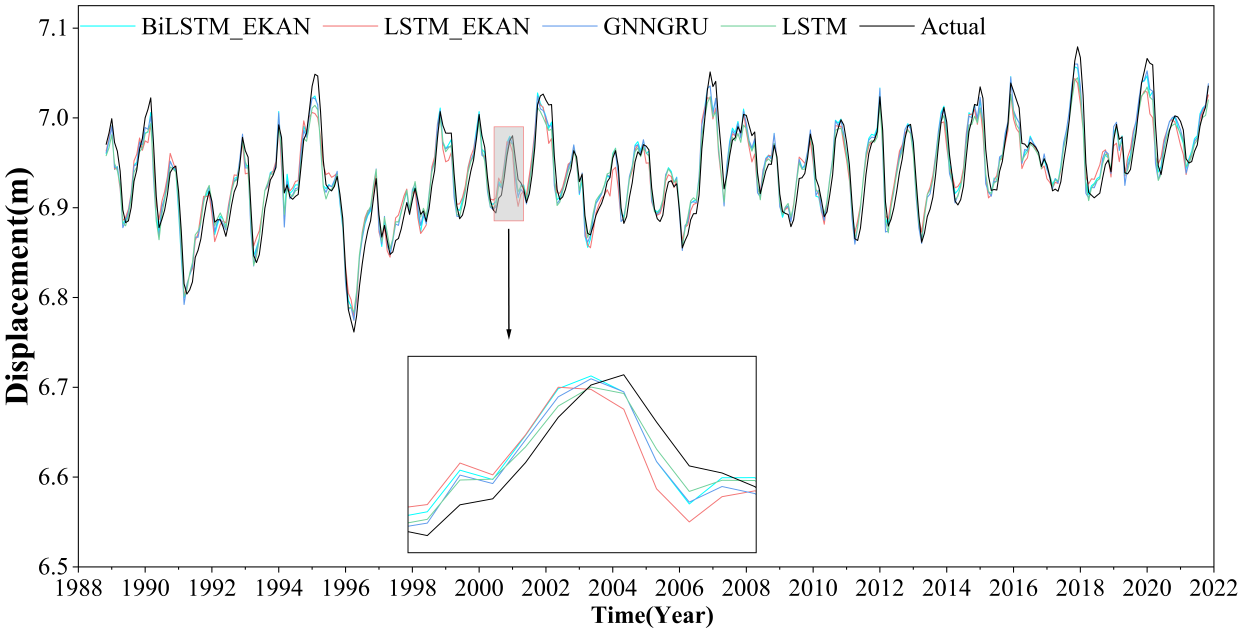}
    \caption{Comparison of LSTM\_ekan, LSTM, GRU and Transformer model fits against true values on the test set.}
    \label{fig:gts_gui}  

\end{figure}
In summary, From the results of Fig.6~8, we can conclude that KAN-optimized models outperform in long-term predictions due to their enhanced ability to capture complex patterns, while simpler MLP-based models (e.g., LSTM, BiLSTM) show better short-term performance by quickly learning key features. Experimental results confirm these theoretical advantages, with KAN variants (e.g., LSTM\_ekan) achieving higher WQE in long-term tasks and MLPs excelling in short-term predictions.
\section{Conclusions}
GTS-Forecaster is an open-source Python toolkit for geodetic time series (GNSS, TG, and SSH) forecasting. It integrates tools for gap-filling, forecasting, visualization, and accuracy assessment of geodetic time series. Experimental results showed that GTS-Forecaster effectively models complex spatiotemporal patterns, including nonlinear dynamics of crustal deformation and sea level variability. In addition, GTS Forecaster also serves as a versatile tool for general time series prediction. Future work will focus on expanding model interoperability with geophysical simulation outputs. In our future work, we aim to enhance the GTS Forecaster software in several key directions. First, we plan to integrate more advanced deep learning models and hybrid architectures to further improve the accuracy and robustness of geodetic time series forecasting. This includes exploring the potential of physics-informed neural networks (PINNs) that can incorporate geophysical laws and constraints directly into the learning process, thereby enhancing the physical plausibility of predictions. Second, we intend to expand the software's capabilities to handle multi-source geodetic data fusion, enabling more comprehensive analysis of geodynamic processes by combining data from GNSS, tide gauges, satellite altimetry, and other emerging sensors like satellite gravity missions. Third, we will focus on optimizing the computational efficiency of the models, particularly for large-scale datasets, by implementing techniques such as model parallelism and more efficient training algorithms. Fourth, we aim to develop advanced visualization tools to help users better interpret the forecasting results and underlying spatiotemporal patterns. Lastly, we will work on improving the user interaction experience by refining the GUI and providing more detailed documentation and tutorials to facilitate wider adoption within the geoscientific community and beyond. Through these efforts, we hope to make GTS Forecaster a more powerful and versatile tool for geodetic time series analysis and forecasting.

\section*{Acknowledgements}
This work was sponsored by National Natural Science Foundation of China (42364002), Major Discipline Academic and Technical Leaders Training Program of Jiangxi Province (20225BCJ23014).

\bibliographystyle{unsrt}  


\begin{thebibliography}{1}

\bibitem{he2017review}
X.~He, J.-P.~Montillet, R.~Fernandes, M.~Bos, K.~Yu, X.~Hua, and W.~Jiang,
``Review of current GPS methodologies for producing accurate time series and their error sources,''
\emph{Journal of Geodynamics}, vol.~106, pp.~12--29, 2017.

\bibitem{blewitt2016midas}
G.~Blewitt, C.~Kreemer, W.~C.~Hammond, and J.~Gazeaux,
``MIDAS robust trend estimator for accurate GPS station velocities without step detection,''
\emph{Journal of Geophysical Research: Solid Earth}, vol.~121, no.~3, pp.~2054--2068, 2016.

\bibitem{fialko2006interseismic}
Y.~Fialko,
``Interseismic strain accumulation and the earthquake potential on the southern San Andreas fault system,''
\emph{Nature}, vol.~441, no.~7096, pp.~968--971, 2006.

\bibitem{avouac2015geodetic}
J.-P.~Avouac,
``From geodetic imaging of seismic and aseismic fault slip to dynamic modeling of the seismic cycle,''
\emph{Annual Review of Earth and Planetary Sciences}, vol.~43, pp.~233--271, 2015.

\bibitem{montillet2019geodetic}
J.-P.~Montillet and M.~S.~Bos,
\emph{Geodetic Time Series Analysis in Earth Sciences}.
Springer, 2019.

\bibitem{rebischung2016igs}
P.~Rebischung, Z.~Altamimi, J.~Ray, and B.~Garayt,
``The IGS contribution to ITRF2014,''
\emph{Journal of Geodesy}, vol.~90, no.~7, pp.~611--630, 2016.

\bibitem{box2015time}
G.~E.~P.~Box, G.~M.~Jenkins, G.~C.~Reinsel, and G.~M.~Ljung,
\emph{Time Series Analysis: Forecasting and Control}, 5th~ed.
John Wiley \& Sons, 2015.

\bibitem{hyndman2018forecasting}
R.~J.~Hyndman and G.~Athanasopoulos,
\emph{Forecasting: Principles and Practice}, 2nd~ed.
OTexts, 2018.

\bibitem{lecun2015deep}
Y.~LeCun, Y.~Bengio, and G.~Hinton,
``Deep learning,''
\emph{Nature}, vol.~521, no.~7553, pp.~436--444, 2015.

\bibitem{goodfellow2016deep}
I.~Goodfellow, Y.~Bengio, and A.~Courville,
\emph{Deep Learning}, vol.~1.
MIT Press, 2016.

\bibitem{li2025modeling}
Z.~Li and T.~Lu,
``Modeling of regional GNSS network using adaptive boosting algorithm: a case study in the Xinjiang Uyghur Autonomous Region,''
\emph{GPS Solutions}, vol.~29, no.~1, 25, 2025.

\bibitem{liu2024kan}
Z.~Liu, Y.~Wang, S.~Vaidya, F.~Ruehle, J.~Halverson, M.~Soljačić, T.~Y.~Hou, and M.~Tegmark,
``KAN: Kolmogorov–Arnold networks,''
arXiv:2404.19756, 2024.

\bibitem{xu2023timegnn}
N.~Xu, C.~Kosma, and M.~Vazirgiannis,
``TimeGNN: temporal dynamic graph learning for time series forecasting,''
in \emph{Int.~Conf.~on Complex Networks and Their Applications}, pp.~87--99, Springer, Cham, 2023.

\bibitem{zhou2025improved}
Y.~Zhou, X.~He, J.-P.~Montillet, S.~Wang, S.~Hu, X.~Sun, J.~Huang, and X.~Ma,
``An improved ICEEMDAN-MPA-GRU model for GNSS height time series prediction with weighted quality evaluation index,''
\emph{GPS Solutions}, 2025.

\bibitem{ackerman2020detection}
S.~Ackerman, E.~Farchi, O.~Raz, M.~Zalmanovici, and P.~Dube,
``Detection of data drift and outliers affecting machine learning model performance over time,''
arXiv:2012.09258, 2020.

\bibitem{seo2006review}
S.~Seo,
\emph{A Review and Comparison of Methods for Detecting Outliers in Univariate Data Sets}.
PhD thesis, University of Pittsburgh, 2006.

\bibitem{arismendi2021piecewise}
R.~Arismendi, A.~Barros, and A.~Grall,
``Piecewise deterministic Markov process for condition-based maintenance models—application to critical infrastructures with discrete-state deterioration,''
\emph{Reliability Engineering \& System Safety}, vol.~212, 107540, 2021.

\bibitem{xie2025stability}
S.~Xie, D.~Gan, and Z.~Liu,
``Stability analysis of distributed Kalman filtering algorithm for stochastic regression model,''
\emph{Control Theory and Technology}, pp.~1--15, 2025.

\bibitem{hochreiter1997long}
S.~Hochreiter and J.~Schmidhuber,
``Long short-term memory,''
\emph{Neural Computation}, vol.~9, no.~8, pp.~1735--1780, 1997.

\bibitem{cho2014learning}
K.~Cho, B.~van Merriënboer, C.~Gulcehre, D.~Bahdanau, F.~Bougares, H.~Schwenk, and Y.~Bengio,
``Learning phrase representations using RNN encoder–decoder for statistical machine translation,''
arXiv:1406.1078, 2014.

\bibitem{bai2018empirical}
S.~Bai, J.~Z.~Kolter, and V.~Koltun,
``An empirical evaluation of generic convolutional and recurrent networks for sequence modeling,''
arXiv:1803.01271, 2018.

\bibitem{girshick2015fast}
R.~Girshick,
``Fast R-CNN,''
in \emph{Proc.~IEEE Int.~Conf.~on Computer Vision}, pp.~1440--1448, 2015.

\bibitem{shi2017deep}
X.~Shi, Z.~Gao, L.~Lausen, H.~Wang, D.-Y.~Yeung, W.-K.~Wong, and W.-C.~Woo,
``Deep learning for precipitation nowcasting: a benchmark and a new model,''
in \emph{Advances in Neural Information Processing Systems}, vol.~30, 2017.

\bibitem{huang2015bidirectional}
Z.~Huang, W.~Xu, and K.~Yu,
``Bidirectional LSTM-CRF models for sequence tagging,''
arXiv:1508.01991, 2015.

\bibitem{zhou2021informer}
H.~Zhou, S.~Zhang, J.~Peng, S.~Zhang, J.~Li, H.~Xiong, and W.~Zhang,
``Informer: beyond efficient transformer for long sequence time-series forecasting,''
in \emph{Proc.~AAAI Conf.~on Artificial Intelligence}, vol.~35, no.~12, pp.~11106--11115, 2021.

\bibitem{rashnu2024integrative}
A.~Rashnu and A.~Salimi-Badr,
``Integrative deep learning framework for Parkinson's disease early detection using gait cycle data measured by wearable sensors: a CNN-GRU-GNN approach,''
arXiv:2404.15335, 2024.

\end{thebibliography}

\end{document}